\documentclass[11pt]{article}
\usepackage{booktabs}
\usepackage{stfloats}
\usepackage{balance}
\usepackage{tikz}
\usetikzlibrary{positioning, arrows.meta, calc, fit, shapes.geometric}
\usepackage[export]{adjustbox}  
\usepackage[most]{tcolorbox}
\newtcbox{\maskbox}[1][]{on line, arc=3pt, outer arc=3pt,
  colback=gray!20, colframe=gray!50, boxsep=0pt,
  left=3pt, right=3pt, top=2pt, bottom=2pt,
  boxrule=0.6pt, fontupper=\small\ttfamily\strut, #1}

\newtcbox{\commitbox}[1][]{on line, arc=3pt, outer arc=3pt,
  colback=blue!8, colframe=blue!40, boxsep=0pt,
  left=3pt, right=3pt, top=2pt, bottom=2pt,
  boxrule=0.6pt, fontupper=\small\ttfamily\strut, #1}
  
  \newtcbox{\promptbox}[1][]{on line, arc=5pt, outer arc=5pt,
  colback=gray!5, colframe=gray!40, boxsep=2pt,
  left=6pt, right=6pt, top=4pt, bottom=4pt,
  boxrule=0.8pt, #1}
\usepackage{amsfonts}
\usepackage{amsmath}
\DeclareMathOperator*{\argmin}{arg\,min}
\usepackage[final]{acl}

\usepackage{times}
\usepackage{latexsym}
\usepackage{algorithm}
\usepackage{algorithmic}
\usepackage{placeins}
\usepackage[T1]{fontenc}

\usepackage[utf8]{inputenc}

\usepackage{microtype}

\usepackage{inconsolata}

\usepackage{graphicx}

\title{Training-Free Character-Length Control \\ in Summarization with Diffusion Language Models}

\author{
  \textbf{Steffen Castle\textsuperscript{1}} \qquad
  \textbf{Nils Feldhus\textsuperscript{3,1}} \qquad
  \textbf{Christopher Ebert\textsuperscript{1}} \\
  \textbf{Leonhard Hennig\textsuperscript{1}} \qquad
  \textbf{Sebastian Möller\textsuperscript{1,2}}
  \\[0.5em]
  \textsuperscript{1}German Research Center for Artificial Intelligence (DFKI) \\
  \textsuperscript{2}Technische Universit\"at Berlin
  \qquad
  \textsuperscript{3}CLCG, University of Groningen
}

\begin{document}
\maketitle
\begin{abstract}
Summarization with a fixed character length is important for tasks such as headline and article preview generation, where text must fit an exact space. Autoregressive language models (LMs) struggle with precise character-length control: training-based methods require task-specific fine-tuning, while zero-shot prompting strategies achieve only approximate compliance. We propose \textsc{Delta} (\textbf{D}iffusion \textbf{E}xpected \textbf{L}ength \textbf{T}argeting \textbf{A}lgorithm), a training-free inference-time method for masked diffusion LMs that estimates expected character length at each denoising step and dynamically inserts or removes masked positions to steer toward a target length. \textsc{Delta} requires no fine-tuning and is designed for off-the-shelf masked diffusion LMs. Across three summarization benchmarks, it reduces character-length errors by an order of magnitude compared to baselines while maintaining competitive summary quality in both automatic and human evaluation.
\end{abstract}
\section{Introduction}

Many summarization tasks require outputs that fit a precise character budget. In publishing workflows, for example, headlines and article previews are often produced for predefined display slots with fixed character capacities \citep{hitomi2019large}; similar constraints arise in notification previews, social media posts, subtitles, and UI text, where even a few extra characters can cause truncation or layout overflow. Existing approaches to length-controlled summarization operate within the constraints of autoregressive generation: training-based methods \citep{miculicich2023summarization, belligoli2025controllingsummarizationlengtheos} modify the model or its objective to encourage a target length, while zero-shot strategies \citep{retkowski2025zeroshot, xie2025promptbasedoneshotexactlengthcontrolled} rely on prompting techniques or post-hoc filtering. Both approaches operate under standard left-to-right autoregressive decoding, where previously emitted tokens are not directly revised within the same generation pass. This makes it difficult to dynamically correct the decoded character length as generation proceeds.

Masked diffusion language models such as LLaDA \citep{nie2025llada} and Dream \citep{ye2025dream} offer a structurally different generation process. These models iteratively refine a buffer of masked tokens over multiple denoising steps, and are trained to generate across a range of output lengths. This means that occasional insertions or removals of mask tokens will not severely disrupt generation quality. We exploit this property with \textsc{Delta}, an inference-time method that estimates expected character length from both committed tokens and the probability distributions over remaining masks, then adjusts the buffer to steer toward a target. \textsc{Delta} requires no fine-tuning or architectural changes and is designed to apply to off-the-shelf masked diffusion LMs. Across three summarization benchmarks, it reduces character-length errors by an order of magnitude compared to all baselines while maintaining competitive summary quality.

\section{Background and Related Work}

\paragraph{Length-Controlled Summarization.}
Prior work on length-controlled summarization predominantly operates within the autoregressive framework and spans token-, word-, sentence-, and character-level constraints. Training-based approaches modify either the model or its objective: CTRLsum \citep{he2020ctrlsum} conditions generation on coarse length bins, \citet{miculicich2023summarization} jointly train a length predictor and constrained decoder for precise token or sentence counts, and \citet{belligoli2025controllingsummarizationlengtheos} upweight the EOS-token loss to enforce a character-length ceiling.

Training-free approaches instead steer pretrained autoregressive LMs at inference time. \citet{retkowski2025zeroshot} convert character targets to word targets using statistical ratios and filter multiple samples, while \citet{xie2025promptbasedoneshotexactlengthcontrolled} append countdown markers to obtain near-exact word-level control. Instruction- and position-based methods likewise target explicit length constraints, including sentence compression with InstructCMP \citep{juseon-do-etal-2024-instructcmp} and positional representations for length control \citep{wang-etal-2024-positionid}. Other inference-time approaches address precise length control in LLMs \citep{butcher2024precise}, black-box length-constrained generation \citep{gu-etal-2025-length}, and iterative controllable summarization \citep{ryu-etal-2026-exploring}.

Controllable generation has also been studied in diffusion language models, including methods that steer generation toward desired attributes \citep{li2022diffusionlm}. In contrast to these approaches, \textsc{Delta} exploits masked-diffusion denoising to directly estimate and dynamically correct decoded character length during generation, requiring neither model training nor post-hoc regeneration.

\paragraph{Discrete Diffusion Language Models.}
D3PM \citep{austin2021d3pm} introduced discrete diffusion over token vocabularies, and MDLM \citep{sahoo2024simple} refined the training objective with a Rao-Blackwellized loss that substantially narrows the likelihood gap with autoregressive models. These foundations enabled scaling to billion parameters: LLaDA \citep{nie2025llada} trains an 8B masked diffusion model from scratch, competitive with LLaMA~3, while Dream \citep{ye2025dream} initializes from a pretrained autoregressive checkpoint with context-adaptive noise rescheduling. Both generate output by iteratively unmasking tokens using confidence-based strategies originating in MaskGIT \citep{chang2022maskgit}.

\paragraph{Variable-Length Generation in Diffusion LMs.}
Standard masked diffusion decoding requires choosing a fixed token-buffer size before denoising. This token-buffer length is distinct from a target decoded character length because tokens correspond to strings of varying character length. DreamOn \citep{wu2026dreamon} introduces special expand/delete tokens that allow the model to adjust the token-buffer length during denoising, requiring fine-tuning. DAEDAL \citep{li2026beyond} and $\rho$-EOS \citep{yang2026rhoeos} are training-free alternatives that adapt sequence length during open-ended generation. These methods address variable-length generation rather than generation toward a specified decoded character budget.

\begin{figure}[ht!]
  \centering
  \includegraphics[width=0.8\columnwidth, trim=130 180 390 50, clip]{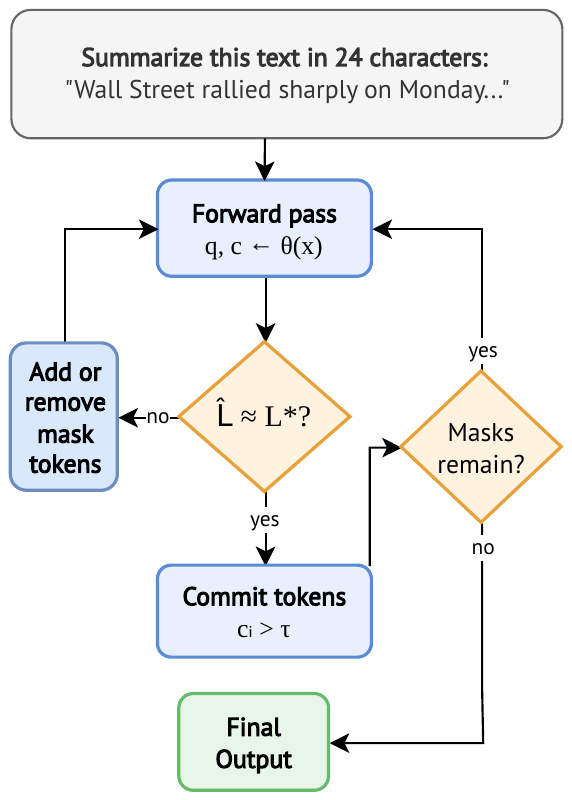}

  \vspace{12pt}

  {\small\sffamily
  \fontfamily{PTSans-TLF}\selectfont
  \textbf{Step 1}\\[3pt]
  \maskbox{MASK}\,\maskbox{MASK}\,\maskbox{MASK}\,\maskbox{MASK}\,\maskbox{MASK}\,\maskbox{MASK}\,\maskbox{MASK}\\[2pt]
  $\hat{L} = 29.4$ \quad {\color{red!70!black}too long $\rightarrow$ remove 1 mask}

  $\downarrow$

  \textbf{Step 2}\\[3pt]
  \maskbox{MASK}\,\maskbox{MASK}\,\maskbox{MASK}\,\maskbox{MASK}\,\maskbox{MASK}\,\maskbox{MASK}\\[2pt]
  $\hat{L} = 24.3$ \quad {\color{blue!70!black}$\approx L^*$ $\rightarrow$ commit 2 tokens}

  $\downarrow$

  \textbf{Step 3}\\[3pt]
  \commitbox{Stocks}\,\commitbox{surged}\,\maskbox{MASK}\,\maskbox{MASK}\,\maskbox{MASK}\,\maskbox{MASK}\\[2pt]
  $\hat{L} = 23.8$ \quad {\color{blue!70!black}$\approx L^*$ $\rightarrow$ commit 4 tokens}

  $\downarrow$

  \textbf{Step 4}\\[3pt]
  \commitbox{Stocks}\,\commitbox{surged}\,\commitbox{on}\,\commitbox{Mon}\,\commitbox{day}\,\commitbox{.}\\[2pt]
  $\hat{L} = 24$ \quad {\color{green!50!black}done\,\checkmark\, (24 characters)}

  \vspace{6pt}
  {\footnotesize
  \maskbox{MASK} = masked \qquad \commitbox{tok} = committed}
  }
  \caption{Overview of \textsc{Delta}. \textbf{Top:} Algorithm flowchart. \textbf{Bottom:} Example with target $L^* = 24$.}
  \label{fig:overview}
  \vspace{-1em}
\end{figure}

\section{Methodology: \textsc{Delta}}

We describe \textsc{Delta} (\textbf{D}iffusion \textbf{E}xpected \textbf{L}ength \textbf{T}argeting \textbf{A}lgorithm), an inference-time method that steers masked diffusion language models toward a target character length. \textsc{Delta} operates within the diffusion iterative denoising loop and requires no fine-tuning or architectural changes.

\subsection{Preliminaries}

Let $V$ denote the vocabulary and $\ell: V \to \mathbb{N}_0$ the character-length function, which maps each token to the number of characters in its decoded string (with $\ell(v) = 0$ for special tokens such as \texttt{[MASK]}, \texttt{[PAD]}, and \texttt{[EOT]}). Because tokens may decode to strings of different character lengths, fixing the number of output tokens does not in general fix the decoded character length. Given a prompt $\mathbf{p} = (p_1, \ldots, p_n)$, a masked diffusion LM generates by initializing an output buffer $\mathbf{x} = (x_1, \ldots, x_m)$ with all positions set to \texttt{[MASK]} and iteratively predicting tokens at masked positions until none remain.

At each denoising step, the model produces a probability distribution $\mathbf{q}_i \in \Delta^{|V|}$ over the vocabulary for each output position $i$. These distributions, combined with the character-length function $\ell$, provide an estimate of the output's character length at every point during generation, even before any tokens are committed.

\subsection{Expected Character Length}

We define the \emph{expected character length} of the current output as
$
  \hat{L} = \sum_{i=1}^{m} \hat{\ell}_i,
$
where $\hat{\ell}_i$ depends on whether position $i$ has been committed (unmasked) or is still masked:
\begin{equation}
  \hat{\ell}_i =
  \begin{cases}
    \ell(x_i) & \text{if } x_i \neq \texttt{[MASK]}, \\
    \sum_{v \in V} q_{i,v} \cdot \ell(v) & \text{if } x_i = \texttt{[MASK]}.
  \end{cases}
\end{equation}
For committed tokens, $\hat{\ell}_i$ is the exact character length. For masked positions, $\hat{\ell}_i$ is the probability-weighted expected character length under the model's current prediction. We additionally mask out zero-length (special) tokens from logits before computing $\mathbf{q}_i$, ensuring they are never predicted.

\subsection{Confidence Scoring}

At each step, the algorithm must decide which masked positions to unmask. Following MaskGIT-style decoding \citep{chang2022maskgit}, we assign a confidence score $c_i$ to each position and unmask those exceeding a threshold $\tau$. We use the maximum predicted probability as the confidence score: $c_i = \max_{v \in V} q_{i,v}$. At each step, at least one position is always unmasked (the one with highest confidence), guaranteeing progress.

\subsection{Length-Controlled Denoising}

The algorithm combines three operations: \emph{length estimation}, \emph{buffer adjustment}, and \emph{token commitment}. A schematic is shown in Algorithm~\ref{alg:main}.

\paragraph{Buffer adjustment.} After computing $\hat{L}$ and the confidence scores, the algorithm compares $\hat{L}$ to the target $L^*$. If $\hat{L} > L^*$, the buffer is too long and masked positions should be removed. We sort masked positions by ascending confidence (least confident first) and greedily remove masks until the expected length deficit is minimized:
\begin{equation}
  n_{\text{remove}} = \argmin_{k} \left| \hat{L} - \sum_{j=1}^{k} \hat{\ell}_{\sigma(j)} - L^* \right|,
\end{equation}
where $\sigma$ is the sorted ordering. Removal is restricted to masks whose confidence falls below a threshold $\tau_{\text{low}}$, preventing deletion of positions the model is already confident about. If instead $\hat{L} < L^*$, new \texttt{[MASK]} tokens are inserted adjacent to the least-confident masked positions. The number of insertions is estimated as $(L^* - \hat{L}) / \bar{\ell}$, where $\bar{\ell}$ is the current mean expected character length per token.

\begin{algorithm}[t]
\caption{Length-Controlled Diffusion Decoding}
\label{alg:main}
\begin{algorithmic}[1]
\REQUIRE Diffusion LM $\theta$, prompt $\mathbf{p}$, target length $L^*$, initial buffer size $m$, confidence threshold $\tau$
\STATE $\mathbf{x} \leftarrow [\mathbf{p}; \texttt{MASK}^m; \texttt{EOT}]$
\STATE $\textit{can\_adjust} \leftarrow \text{true}$
\WHILE{$\mathbf{x}$ contains \texttt{MASK} tokens}
    \STATE $\mathbf{q}, \mathbf{c} \leftarrow \text{ForwardPass}(\theta, \mathbf{x})$
    \STATE $\hat{L} \leftarrow \text{ExpectedCharLength}(\mathbf{x}, \mathbf{q})$
    \IF{$\hat{L} > L^*$ \AND $\textit{can\_adjust}$}
        \STATE Remove lowest-confidence masks to reduce $\hat{L}$ toward $L^*$
        \STATE $\textit{can\_adjust} \leftarrow \text{false}$
    \ELSIF{$\hat{L} < L^*$ \AND $\textit{can\_adjust}$}
        \STATE Insert new masks at lowest-confidence positions
        \STATE $\textit{can\_adjust} \leftarrow \text{false}$
    \ELSE
        \STATE Unmask positions where $c_i > \tau$ (at least one)
        \STATE $\textit{can\_adjust} \leftarrow \text{true}$
    \ENDIF
\ENDWHILE
\RETURN $\mathbf{x}$
\end{algorithmic}
\end{algorithm}

\paragraph{Token commitment.} After any buffer adjustment, masked positions with confidence $c_i > \tau$ are unmasked by replacing them with their argmax predictions. To prevent oscillation, where the algorithm alternates between inserting and removing masks without making progress, buffer adjustments are only permitted immediately after a commitment step. Since each commitment step unmasks at least one token, this guarantees that the sequence state changes between consecutive adjustments, breaking potential cycles. The algorithm terminates when no masked positions remain or a maximum iteration count is reached.

\section{Evaluation}

\begin{table*}[!ht]
  \centering
  \small
  \renewcommand*{\arraystretch}{1.05}
  \begin{tabular*}{\textwidth}{@{\extracolsep{\fill}}l|cc|ccc|cccc@{}}    \toprule
    & \multicolumn{2}{c|}{\textbf{Length}} & \multicolumn{3}{c|}{\textbf{Quality}} & \multicolumn{4}{c}{\textbf{G-Eval}} \\
    \textbf{Method} & \textbf{MAE} & \textbf{$<$10} & \textbf{R-1} & \textbf{R-L} & \textbf{BF1} & \textbf{Rel} & \textbf{Con} & \textbf{Coh} & \textbf{Flu} \\
    \midrule
    \multicolumn{10}{l}{XSum} \\
    \midrule
    \quad\textsc{Delta}+LLaDA & \textbf{2.3} & 99.6 & 0.286 & 0.217 & 0.882 & 3.41 & 4.21 & 4.61 & 4.66 \\
    \quad\textsc{Delta}+Dream & \textbf{2.3} & \textbf{99.7} & \textbf{0.333} & \textbf{0.255} & \textbf{0.890} & 3.17 & 4.14 & 4.76 & 4.82 \\
    \quad LA-SF-AR & 5.2 & 84.2 & 0.273 & 0.204 & 0.881 & 3.56 & 4.38 & 4.79 & 4.89 \\
    \quad GPT-4o-mini & 13.1 & 48.4 & 0.264 & 0.194 & 0.880 & \textbf{3.83} & 4.73 & \textbf{4.96} & \textbf{4.99} \\
    \quad Lead-char-snap & 15.7 & 60.5 & 0.166 & 0.122 & 0.850 & 1.80 & 4.72 & 3.08 & 3.31 \\
    \quad TextRank & 14.5 & 50.1 & 0.178 & 0.130 & 0.845 & 1.74 & \textbf{4.79} & 4.07 & 4.59 \\
    \midrule
    \multicolumn{10}{l}{CNN/DailyMail} \\
    \midrule
    \quad\textsc{Delta}+LLaDA & \textbf{2.4} & \textbf{99.6} & 0.363 & 0.259 & 0.876 & 4.04 & 4.59 & 4.58 & 4.49 \\
    \quad\textsc{Delta}+Dream & 3.0 & 99.3 & 0.409 & 0.281 & 0.880 & 3.97 & 4.72 & 4.75 & 4.73 \\
    \quad LA-SF-AR & 10.3 & 59.7 & 0.399 & 0.254 & 0.879 & 4.29 & 4.61 & 4.94 & 4.96 \\
    \quad GPT-4o-mini & 96.6 & 11.1 & 0.393 & 0.245 & 0.877 & \textbf{4.53} & 4.86 & \textbf{4.99} & \textbf{5.00} \\
    \quad BART-CNN & 104.7 & 5.7 & \textbf{0.441} & \textbf{0.306} & \textbf{0.882} & 3.69 & 4.83 & 4.60 & 4.61 \\
    \quad Lead-char-snap & 59.4 & 21.4 & 0.363 & 0.230 & 0.868 & 2.85 & \textbf{4.92} & 4.22 & 4.40 \\
    \quad TextRank & 17.3 & 32.4 & 0.356 & 0.235 & 0.866 & 2.40 & 4.80 & 3.38 & 4.07 \\
    \midrule
    \multicolumn{10}{l}{arXiv} \\
    \midrule
    \quad\textsc{Delta}+LLaDA & \textbf{2.8} & \textbf{98.4} & 0.443 & 0.234 & 0.837 & 3.71 & 4.35 & 4.43 & 4.39 \\
    \quad\textsc{Delta}+Dream & 3.1 & 98.3 & 0.431 & 0.236 & \textbf{0.853} & 3.09 & 4.42 & 3.60 & 3.78 \\
    \quad LA-SF-AR & 44.9 & 22.9 & \textbf{0.466} & \textbf{0.250} & 0.845 & 3.98 & 4.28 & 4.78 & 4.98 \\
    \quad GPT-4o-mini & 112.2 & 3.7 & 0.432 & 0.217 & 0.839 & \textbf{4.35} & 4.65 & \textbf{4.97} & \textbf{5.00} \\
    \quad Lead-char-snap & 96.5 & 7.0 & 0.351 & 0.183 & 0.837 & 2.22 & \textbf{4.99} & 4.11 & 4.27 \\
    \quad TextRank & 46.9 & 12.0 & 0.406 & 0.212 & 0.837 & 2.52 & 4.55 & 3.00 & 3.19 \\    
    \bottomrule
  \end{tabular*}
\caption{Results on XSum, CNN/DailyMail, and arXiv, ordered by increasing reference summary length. MAE is mean absolute character-length error; $<$10 is the percentage of outputs within 10 characters of the target. R-1/R-L = ROUGE-1/L; BF1 = BERTScore F1. G-Eval scores (1--5 Likert) use GPT-5-mini as judge \citep{liu2023geval}. Best values per column within each dataset are \textbf{bolded}. BART-CNN is included only for CNN/DailyMail as it was fine-tuned on that dataset.}
\label{tab:results}
  \vspace*{-1em}
\end{table*}

We evaluate on three datasets: CNN/DailyMail \citep{hermann2015cnndm}, a news summarization benchmark; XSum \citep{narayan2018xsum}, which targets single-sentence summaries; and arXiv \citep{cohan2018discourse}, which contains scientific paper abstracts. For each reference summary, we set the target character length to the character length of the reference. Dataset statistics are provided in Appendix~\ref{sec:dataset_stats}.

\paragraph{Baselines.} We compare \textsc{Delta} with LLaDA-8B-Instruct \citep{nie2025llada} and Dream-7B-Instruct \citep{ye2025dream} against: TextRank \citep{mihalcea2004textrank}; Lead-char-snap, which selects the maximum number of initial source sentences that fit within the target character budget; BART fine-tuned on CNN/DailyMail \citep{lewis2020bart}; GPT-4o-mini prompted with the target character count; and LA-SF-AR \citep{retkowski2025zeroshot} using Llama-3.1-8B-Instruct. Full baseline and hyperparameter details are in Appendix~\ref{sec:baseline_details}. We additionally evaluate DAEDAL \citep{li2026beyond}, a training-free variable-length diffusion decoding method, using the same character-target prompts as the other generative methods. Results are reported in Appendix~\ref{sec:daedal}. To isolate the contribution of dynamic buffer adjustment, we also evaluate a fixed-buffer ablation using the same diffusion models, prompts, confidence schedules, and initialization as \textsc{Delta}, but with mask insertion and removal disabled. Results are reported in Appendix~\ref{sec:fixed_buffer}. \footnote{Implementation, evaluation scripts, and experimental configuration are publicly available at \url{https://github.com/DFKI-NLP/delta-diffusion}.}

\paragraph{Metrics.} We report mean absolute character-length error (MAE) and the percentage of outputs within 10 characters of the target, along with summary quality via ROUGE-1, ROUGE-L \citep{lin2004rouge}, and BERTScore F1 \citep{zhang2020bertscore}. We additionally evaluate using LLM-as-a-judge following the G-Eval framework \citep{liu2023geval}, which scores summaries on \emph{relevance} (Rel), \emph{consistency} (Con), \emph{coherence} (Coh), and \emph{fluency} (Flu) on a 1--5 Likert scale. We use GPT-5-mini as the judge; we additionally report GPT-4o-mini scores in the Appendix. All prompts are listed in Appendix~\ref{sec:prompts}.

\paragraph{Human evaluation.}
We additionally conduct a blind human evaluation on 50 CNN/DailyMail examples comparing \textsc{Delta}+Dream, LA-SF-AR, and GPT-4o-mini. Two annotators independently rate anonymized and randomly ordered outputs on relevance, consistency, coherence, and fluency using the same 1--5 scale as the G-Eval evaluation. Full results and agreement statistics are reported in Appendix~\ref{sec:human_eval}.

\paragraph{Runtime.}
We report wall-clock inference time per summary for \textsc{Delta} and the non-API baselines. Values are measured as mean $\pm$ standard deviation over generated summaries; full results are reported in Appendix~\ref{sec:runtime}.

\section{Results}

Results are shown in Table~\ref{tab:results}. On all datasets, \textsc{Delta} achieves the best length compliance by a wide margin. While baseline errors scale with summary length, \textsc{Delta} maintains MAE below 3.1 characters even on arXiv, where references average 969 characters. LA-SF-AR, the strongest autoregressive baseline for length control, places 84.2\% of outputs within 10 characters on XSum but only 22.9\% on arXiv, whereas \textsc{Delta} remains above 98\% across all datasets. 

On summary quality, \textsc{Delta} remains competitive. On XSum, \textsc{Delta}+Dream achieves the highest ROUGE and BERTScore values, while on arXiv it achieves the highest BERTScore. On CNN/DailyMail, \textsc{Delta}+Dream is second only to the fine-tuned BART baseline on ROUGE and BERTScore. Autoregressive baselines generally receive higher GPT-5-mini G-Eval scores, but this advantage comes with substantially poorer character-length compliance. A complementary human evaluation on 50 CNN/DailyMail examples finds \textsc{Delta}+Dream competitive with LA-SF-AR and GPT-4o-mini, with average ratings of 4.27, 4.32, and 4.31 respectively and 91.5\% within-one-point annotator agreement; full results are reported in Appendix~\ref{sec:human_eval}. Additional analyses confirm that \textsc{Delta}'s length-control gains come from dynamic buffer adjustment. Disabling mask insertion and removal in the fixed-buffer ablation substantially worsens character-length compliance while leaving summary quality broadly similar. Likewise, DAEDAL produces reasonable summaries but does not reliably satisfy specified character budgets. Full results are reported in Appendices~\ref{sec:fixed_buffer} and~\ref{sec:daedal}.

\section{Conclusion}

We presented \textsc{Delta}, a training-free inference-time method that steers masked diffusion language models toward exact character-length targets by estimating expected length from token distributions and dynamically adjusting the generation buffer. Applied to LLaDA and Dream, \textsc{Delta} reduces character-length errors by an order of magnitude compared to strong baselines while maintaining competitive summary quality in both automatic and human evaluation. Future work could address inference speed, evaluate \textsc{Delta} across languages and model scales, and extend the approach to other constrained generation tasks.

\section*{Limitations}

\textsc{Delta} has several limitations. First, inference is slow. Diffusion language models require multiple full-sequence forward passes during denoising and, unlike autoregressive models, cannot use standard KV caching due to their bidirectional attention \citep{nie2025llada}. On XSum, \textsc{Delta} requires 0.67--0.69 seconds per summary, compared with 1.43 seconds for LA-SF-AR. On CNN/DailyMail and arXiv, however, \textsc{Delta} requires 8.59 and 20.75--22.80 seconds per summary, respectively, compared with 2.19 and 13.22 seconds for LA-SF-AR. Thus, the tighter character-length control comes with increased inference cost on longer inputs. Full runtime measurements are reported in Appendix~\ref{sec:runtime}. Recent work on accelerating diffusion LM inference through approximate KV caching and guided decoding \citep{hu2025flashdlmacceleratingdiffusionlanguage} may mitigate this cost.

Second, automatic quality evaluation remains sensitive to the choice of LLM judge. GPT-4o-mini and GPT-5-mini produce substantially different Consistency and Coherence scores for several methods, particularly extractive baselines, consistent with known limitations of LLM-based judges \citep{chehbouni2025neither}. Our human evaluation provides additional evidence that \textsc{Delta}'s length control does not cause a substantial quality regression, but it is limited to 50 CNN/DailyMail examples and two annotators. Larger human evaluations across datasets would provide stronger evidence.

Third, our evaluation is limited to English-language summarization and to two diffusion models in the 7--8B parameter range. Both LLaDA and Dream are predominantly trained on English data and perform poorly on other languages in our informal testing. The effectiveness of character-level length estimation also depends on the token-to-character mapping of the underlying tokenizer, which varies across languages and writing systems. Although \textsc{Delta} operates on output distributions rather than model-specific internal components, its effectiveness on substantially smaller or larger diffusion models remains to be validated.

Finally, the hyperparameters $\tau$ and $\tau_{\text{low}}$ vary across datasets and models, and we selected them via grid search on validation sets. We do not provide a sensitivity analysis or adaptive method for setting these thresholds, which limits the plug-and-play applicability of \textsc{Delta} to new domains without tuning.

\section*{Acknowledgments}

This work was funded by the German Federal Ministry of Research, Technology and Space (BMFTR) under the GenKI4Media project (grant 01MK25004E).

\bibliography{custom}

@inproceedings{hitomi2019large,
    title = "A Large-Scale Multi-Length Headline Corpus for Analyzing Length-Constrained Headline Generation Model Evaluation",
    author = "Hitomi, Yuta  and
      Taguchi, Yuya  and
      Tamori, Hideaki  and
      Kikuta, Ko  and
      Nishitoba, Jiro  and
      Okazaki, Naoaki  and
      Inui, Kentaro  and
      Okumura, Manabu",
    editor = "van Deemter, Kees  and
      Lin, Chenghua  and
      Takamura, Hiroya",
    booktitle = "Proceedings of the 12th International Conference on Natural Language Generation",
    month = oct # "–" # nov,
    year = "2019",
    address = "Tokyo, Japan",
    publisher = "Association for Computational Linguistics",
    url = "https://aclanthology.org/W19-8641/",
    doi = "10.18653/v1/W19-8641",
    pages = "333--343"
}

@inproceedings{li2022diffusionlm,
    author = {Li, Xiang Lisa and Thickstun, John and Gulrajani, Ishaan and Liang, Percy and Hashimoto, Tatsunori B.},
    title = {Diffusion-LM improves controllable text generation},
    year = {2022},
    isbn = {9781713871088},
    publisher = {Curran Associates Inc.},
    address = {Red Hook, NY, USA},
    booktitle = {Proceedings of the 36th International Conference on Neural Information Processing Systems},
    articleno = {313},
    numpages = {16},
    location = {New Orleans, LA, USA},
    series = {NIPS '22}
}

@inproceedings{wang-etal-2024-positionid,
    title = "{P}osition{ID}: {LLM}s can Control Lengths, Copy and Paste with Explicit Positional Awareness",
    author = "Wang, Noah  and
      Duan, Feiyu  and
      Zhang, Yibo  and
      Zhou, Wangchunshu  and
      Xu, Ke  and
      Huang, Wenhao  and
      Fu, Jie",
    editor = "Al-Onaizan, Yaser  and
      Bansal, Mohit  and
      Chen, Yun-Nung",
    booktitle = "Findings of the Association for Computational Linguistics: EMNLP 2024",
    month = nov,
    year = "2024",
    address = "Miami, Florida, USA",
    publisher = "Association for Computational Linguistics",
    url = "https://aclanthology.org/2024.findings-emnlp.983/",
    doi = "10.18653/v1/2024.findings-emnlp.983",
    pages = "16877--16915"
}

@inproceedings{juseon-do-etal-2024-instructcmp,
    title = "{I}nstruct{CMP}: Length Control in Sentence Compression through Instruction-based Large Language Models",
    author = "Juseon-Do  and
      Kwon, Jingun  and
      Kamigaito, Hidetaka  and
      Okumura, Manabu",
    editor = "Ku, Lun-Wei  and
      Martins, Andre  and
      Srikumar, Vivek",
    booktitle = "Findings of the Association for Computational Linguistics: ACL 2024",
    month = aug,
    year = "2024",
    address = "Bangkok, Thailand",
    publisher = "Association for Computational Linguistics",
    url = "https://aclanthology.org/2024.findings-acl.532/",
    doi = "10.18653/v1/2024.findings-acl.532",
    pages = "8980--8996"
}

@article{butcher2024precise,
    author = {Butcher, Bradley and O’Keefe, Michael and Titchener, James},
    year = {2025},
    month = {04},
    pages = {100143},
    title = {Precise length control for large language models},
    volume = {11},
    journal = {Natural Language Processing Journal},
    doi = {10.1016/j.nlp.2025.100143}
}

@inproceedings{gu-etal-2025-length,
    title = "Length Controlled Generation for Black-box {LLM}s",
    author = "Gu, Yuxuan  and
      Wang, Wenjie  and
      Feng, Xiaocheng  and
      Zhong, Weihong  and
      Zhu, Kun  and
      Huang, Lei  and
      Liu, Ting  and
      Qin, Bing  and
      Chua, Tat-Seng",
    editor = "Che, Wanxiang  and
      Nabende, Joyce  and
      Shutova, Ekaterina  and
      Pilehvar, Mohammad Taher",
    booktitle = "Proceedings of the 63rd Annual Meeting of the Association for Computational Linguistics (Volume 1: Long Papers)",
    month = jul,
    year = "2025",
    address = "Vienna, Austria",
    publisher = "Association for Computational Linguistics",
    url = "https://aclanthology.org/2025.acl-long.825/",
    doi = "10.18653/v1/2025.acl-long.825",
    pages = "16878--16895",
    ISBN = "979-8-89176-251-0"
}

@inproceedings{ryu-etal-2026-exploring,
    title = "Exploring Iterative Controllable Summarization with Large Language Models",
    author = "Ryu, Sangwon  and
      Do, Heejin  and
      Kim, Daehui  and
      Yu, Hwanjo  and
      Kim, Dongwoo  and
      Kim, Yunsu  and
      Lee, Gary  and
      Ok, Jungseul",
    editor = "Demberg, Vera  and
      Inui, Kentaro  and
      Marquez, Llu{\'i}s",
    booktitle = "Findings of the {A}ssociation for {C}omputational {L}inguistics: {EACL} 2026",
    month = mar,
    year = "2026",
    address = "Rabat, Morocco",
    publisher = "Association for Computational Linguistics",
    url = "https://aclanthology.org/2026.findings-eacl.26/",
    doi = "10.18653/v1/2026.findings-eacl.26",
    pages = "512--528",
    ISBN = "979-8-89176-386-9"
}

@inproceedings{
    chehbouni2025neither,
    title={Neither Valid nor Reliable? Investigating the Use of {LLM}s as Judges},
    author={Khaoula Chehbouni and Mohammed Haddou and Jackie CK Cheung and Golnoosh Farnadi},
    booktitle={The Thirty-Ninth Annual Conference on Neural Information Processing Systems Position Paper Track},
    year={2025},
    url={https://openreview.net/forum?id=yqKfMr0yvY}
}

@article{hu2025flashdlmacceleratingdiffusionlanguage,
    title={FlashDLM: Accelerating Diffusion Language Model Inference via Efficient KV Caching and Guided Diffusion},
    author={Hu, Zhanqiu and Meng, Jian and Akhauri, Yash and Abdelfattah, Mohamed S. and Seo, Jae-sun and Zhang, Zhiru and Gupta, Udit},
    journal={arXiv preprint arXiv:2505.21467v2},
    year={2025},
    url={https://arxiv.org/abs/2505.21467v2}
}

@misc{miculicich2023summarization,
    title={Summarization with Precise Length Control}, 
    author={Lesly Miculicich and Yujia Xie and Song Wang and Pengcheng He},
    year={2023},
    eprint={2305.05171},
    archivePrefix={arXiv},
    primaryClass={cs.CL},
    url={https://arxiv.org/abs/2305.05171}, 
}

@article{he2020ctrlsum,
    title={CTRLsum: Towards Generic Controllable Text Summarization},
    author={He, Junxian and Kry{\'s}ci{\'n}ski, Wojciech and McCann, Bryan and Rajani, Nazneen and Xiong, Caiming},
    journal={arXiv},
    year={2020}
}

@inproceedings{lewis2020bart,
    title = "{BART}: Denoising Sequence-to-Sequence Pre-training for Natural Language Generation, Translation, and Comprehension",
    author = "Lewis, Mike  and
      Liu, Yinhan  and
      Goyal, Naman  and
      Ghazvininejad, Marjan  and
      Mohamed, Abdelrahman  and
      Levy, Omer  and
      Stoyanov, Veselin  and
      Zettlemoyer, Luke",
    editor = "Jurafsky, Dan  and
      Chai, Joyce  and
      Schluter, Natalie  and
      Tetreault, Joel",
    booktitle = "Proceedings of the 58th Annual Meeting of the Association for Computational Linguistics",
    month = jul,
    year = "2020",
    address = "Online",
    publisher = "Association for Computational Linguistics",
    url = "https://aclanthology.org/2020.acl-main.703/",
    doi = "10.18653/v1/2020.acl-main.703",
    pages = "7871--7880"
}

@inproceedings{mihalcea2004textrank,
    title = "{T}ext{R}ank: Bringing Order into Text",
    author = "Mihalcea, Rada  and
      Tarau, Paul",
    editor = "Lin, Dekang  and
      Wu, Dekai",
    booktitle = "Proceedings of the 2004 Conference on Empirical Methods in Natural Language Processing",
    month = jul,
    year = "2004",
    address = "Barcelona, Spain",
    publisher = "Association for Computational Linguistics",
    url = "https://aclanthology.org/W04-3252/",
    pages = "404--411"
}

@inproceedings{austin2021d3pm,
    author = {Austin, Jacob and Johnson, Daniel D. and Ho, Jonathan and Tarlow, Daniel and van den Berg, Rianne},
    title = {Structured denoising diffusion models in discrete state-spaces},
    year = {2021},
    isbn = {9781713845393},
    publisher = {Curran Associates Inc.},
    address = {Red Hook, NY, USA},
    booktitle = {Proceedings of the 35th International Conference on Neural Information Processing Systems},
    articleno = {1376},
    numpages = {13},
    series = {NIPS '21}
}

@inproceedings{
    sahoo2024simple,
    title={Simple and Effective Masked Diffusion Language Models},
    author={Subham Sekhar Sahoo and Marianne Arriola and Aaron Gokaslan and Edgar Mariano Marroquin and Alexander M Rush and Yair Schiff and Justin T Chiu and Volodymyr Kuleshov},
    booktitle={The Thirty-eighth Annual Conference on Neural Information Processing Systems},
    year={2024},
    url={https://openreview.net/forum?id=L4uaAR4ArM}
}

@inproceedings{nie2025llada,
    title={Large Language Diffusion Models},
    author={Shen Nie and Fengqi Zhu and Zebin You and Xiaolu Zhang and Jingyang Ou and Jun Hu and JUN ZHOU and Yankai Lin and Ji-Rong Wen and Chongxuan Li},
    booktitle={The Thirty-ninth Annual Conference on Neural Information Processing Systems},
    year={2025},
    url={https://openreview.net/forum?id=KnqiC0znVF}
}

@article{ye2025dream,
    title={Dream 7B: Diffusion Large Language Models},
    author={Ye, Jiacheng and Xie, Zhihui and Zheng, Lin and Gao, Jiahui and Wu, Zirui and Jiang, Xin and Li, Zhenguo and Kong, Lingpeng},
    journal={arXiv preprint arXiv:2508.15487},
    year={2025}
}

@InProceedings{chang2022maskgit,
    title = {MaskGIT: Masked Generative Image Transformer},
    author={Huiwen Chang and Han Zhang and Lu Jiang and Ce Liu and William T. Freeman},
    booktitle = {The IEEE Conference on Computer Vision and Pattern Recognition (CVPR)},
    month = {June},
    year = {2022}
}

@inproceedings{wu2026dreamon,
    title={DreamOn: Diffusion Language Models For Code Infilling Beyond Fixed-size Canvas},
    author={Zirui Wu and Lin Zheng and Zhihui Xie and Jiacheng Ye and Jiahui Gao and Shansan Gong and Yansong Feng and Zhenguo Li and Wei Bi and Guorui Zhou and Lingpeng Kong},
    booktitle={The Fourteenth International Conference on Learning Representations},
    year={2026},
    url={https://openreview.net/forum?id=EQTPmqukiU}
}

@inproceedings{li2026beyond,
    title={Beyond Fixed: Training-Free Variable-Length Denoising for Diffusion Large Language Models},
    author={Jinsong Li and Xiaoyi Dong and Yuhang Zang and Yuhang Cao and Jiaqi Wang and Dahua Lin},
    booktitle={The Fourteenth International Conference on Learning Representations},
    year={2026},
    url={https://openreview.net/forum?id=Ic2A2gCseC}
}

@article{yang2026rhoeos,
    title={$\rho$-$\texttt{EOS}$: Training-free Bidirectional Variable-Length Control for Masked Diffusion LLMs},
    author={Yang, Jingyi and Jiang, Yuxian and Shao, Jing},
    journal={arXiv preprint arXiv:2601.22527},
    year={2026}
}

@misc{belligoli2025controllingsummarizationlengtheos,
      title={Controlling Summarization Length Through EOS Token Weighting}, 
      author={Zeno Belligoli and Emmanouil Stergiadis and Eran Fainman and Ilya Gusev},
      year={2025},
      eprint={2506.05017},
      archivePrefix={arXiv},
      primaryClass={cs.CL},
      url={https://arxiv.org/abs/2506.05017}, 
}

@inproceedings{retkowski2025zeroshot,
    title = "Zero-Shot Strategies for Length-Controllable Summarization",
    author = "Retkowski, Fabian  and
      Waibel, Alexander",
    editor = "Chiruzzo, Luis  and
      Ritter, Alan  and
      Wang, Lu",
    booktitle = "Findings of the Association for Computational Linguistics: NAACL 2025",
    month = apr,
    year = "2025",
    address = "Albuquerque, New Mexico",
    publisher = "Association for Computational Linguistics",
    url = "https://aclanthology.org/2025.findings-naacl.34/",
    doi = "10.18653/v1/2025.findings-naacl.34",
    pages = "551--572",
    ISBN = "979-8-89176-195-7"
}

@misc{xie2025promptbasedoneshotexactlengthcontrolled,
      title={Prompt-Based One-Shot Exact Length-Controlled Generation with LLMs}, 
      author = {Xie, Juncheng and Lee, Hung-yi},
      year={2025},
      eprint={2508.13805},
      archivePrefix={arXiv},
      primaryClass={cs.CL},
      url={https://arxiv.org/abs/2508.13805}, 
}

@inproceedings{hermann2015cnndm,
    author = {Hermann, Karl Moritz and Ko{\v c}isk{\'y}, Tom{\'a}{\v s} and Grefenstette, Edward and Espeholt, Lasse and Kay, Will and Suleyman, Mustafa and Blunsom, Phil},
    title = {Teaching machines to read and comprehend},
    year = {2015},
    publisher = {MIT Press},
    address = {Cambridge, MA, USA},
    booktitle = {Proceedings of the 29th International Conference on Neural Information Processing Systems - Volume 1},
    pages = {1693–1701},
    numpages = {9},
    location = {Montreal, Canada},
    series = {NIPS'15}
}

@inproceedings{narayan2018xsum,
    title = "Don{'}t Give Me the Details, Just the Summary! Topic-Aware Convolutional Neural Networks for Extreme Summarization",
    author = "Narayan, Shashi  and
      Cohen, Shay B.  and
      Lapata, Mirella",
    editor = "Riloff, Ellen  and
      Chiang, David  and
      Hockenmaier, Julia  and
      Tsujii, Jun{'}ichi",
    booktitle = "Proceedings of the 2018 Conference on Empirical Methods in Natural Language Processing",
    month = oct # "-" # nov,
    year = "2018",
    address = "Brussels, Belgium",
    publisher = "Association for Computational Linguistics",
    url = "https://aclanthology.org/D18-1206/",
    doi = "10.18653/v1/D18-1206",
    pages = "1797--1807"
}

@inproceedings{cohan2018discourse,
    title = "A Discourse-Aware Attention Model for Abstractive Summarization of Long Documents",
    author = "Cohan, Arman  and
      Dernoncourt, Franck  and
      Kim, Doo Soon  and
      Bui, Trung  and
      Kim, Seokhwan  and
      Chang, Walter  and
      Goharian, Nazli",
    editor = "Walker, Marilyn  and
      Ji, Heng  and
      Stent, Amanda",
    booktitle = "Proceedings of the 2018 Conference of the North {A}merican Chapter of the Association for Computational Linguistics: Human Language Technologies, Volume 2 (Short Papers)",
    month = jun,
    year = "2018",
    address = "New Orleans, Louisiana",
    publisher = "Association for Computational Linguistics",
    url = "https://aclanthology.org/N18-2097/",
    doi = "10.18653/v1/N18-2097",
    pages = "615--621"
}

@inproceedings{lin2004rouge,
    title = "{ROUGE}: A Package for Automatic Evaluation of Summaries",
    author = "Lin, Chin-Yew",
    booktitle = "Text Summarization Branches Out",
    month = jul,
    year = "2004",
    address = "Barcelona, Spain",
    publisher = "Association for Computational Linguistics",
    url = "https://aclanthology.org/W04-1013/",
    pages = "74--81"
}

@inproceedings{zhang2020bertscore,
    title={BERTScore: Evaluating Text Generation with BERT},
    author={Tianyi Zhang and Varsha Kishore and Felix Wu and Kilian Q. Weinberger and Yoav Artzi},
    booktitle={International Conference on Learning Representations},
    year={2020},
    url={https://openreview.net/forum?id=SkeHuCVFDr}
}

@inproceedings{liu2023geval,
    title = "{G}-Eval: {NLG} Evaluation using Gpt-4 with Better Human Alignment",
    author = "Liu, Yang  and
      Iter, Dan  and
      Xu, Yichong  and
      Wang, Shuohang  and
      Xu, Ruochen  and
      Zhu, Chenguang",
    editor = "Bouamor, Houda  and
      Pino, Juan  and
      Bali, Kalika",
    booktitle = "Proceedings of the 2023 Conference on Empirical Methods in Natural Language Processing",
    month = dec,
    year = "2023",
    address = "Singapore",
    publisher = "Association for Computational Linguistics",
    url = "https://aclanthology.org/2023.emnlp-main.153/",
    doi = "10.18653/v1/2023.emnlp-main.153",
    pages = "2511--2522"
}

\appendix
\raggedbottom

\section{Dataset Statistics}
\label{sec:dataset_stats}

\begin{table}[h]
  \centering
  \small
  \begin{tabular}{lrrr}
    \toprule
    & \textbf{XSum} & \textbf{CNN/DM} & \textbf{arXiv} \\
    \midrule
    Test examples & 11{,}334 & 11{,}490 & 6{,}440 \\
    Avg src length (words) & 376 & 684 & 5{,}906 \\
    Avg ref length (words) & 21 & 55 & 163 \\
    Avg ref length (chars) & 125 & 312 & 969 \\
    \bottomrule
  \end{tabular}
  \caption{Dataset statistics for the test splits used in our evaluation.}
  \label{tab:dataset_stats}
\end{table}

\section{Human Evaluation}
\label{sec:human_eval}

We conduct a blind human evaluation on 50 CNN/DailyMail examples comparing \textsc{Delta}+Dream, LA-SF-AR, and GPT-4o-mini. Two annotators independently rated anonymized and randomly ordered outputs on relevance, consistency, coherence, and fluency using a 1--5 Likert scale.

\begin{table}[h]
  \centering
  \small
  \begin{tabular}{lccccc}
    \toprule
    \textbf{Method} & \textbf{Rel} & \textbf{Con} & \textbf{Coh} & \textbf{Flu} & \textbf{Avg.} \\
    \midrule
    \textsc{Delta}+Dream & 4.05 & 4.33 & 4.31 & 4.40 & 4.27 \\
    LA-SF-AR & 4.18 & 4.36 & 4.32 & 4.42 & 4.32 \\
    GPT-4o-mini & 4.26 & 4.32 & 4.26 & 4.42 & 4.31 \\
    \bottomrule
  \end{tabular}
  \caption{Blind two-annotator human evaluation on 50 CNN/DailyMail examples. Within-one-point annotator agreement is 91.5\% across dimensions.}
  \label{tab:human_eval}
\end{table}

All three methods receive similar ratings across the four dimensions, providing additional evidence that \textsc{Delta}'s substantially tighter character-length control does not come at the cost of a substantial degradation in perceived summary quality.

\section{Fixed-Buffer Ablation}
\label{sec:fixed_buffer}

To isolate the effect of \textsc{Delta}'s dynamic buffer adjustment, we compare against fixed-buffer diffusion decoding using the same model, prompt, confidence schedule, and initialization as \textsc{Delta}, but with mask insertion and removal disabled. As shown in Table~\ref{tab:fixed_buffer}, disabling mask insertion and removal substantially reduces character-length compliance while leaving summary quality broadly similar.

\begin{table*}[h]
  \centering
  \small
  \begin{tabular}{llrrrrrrrrr}
    \toprule
    \textbf{Dataset} & \textbf{Method} & \textbf{MAE} & \textbf{$<$10} &
    \textbf{R-1} & \textbf{R-L} & \textbf{BF1} &
    \textbf{Rel} & \textbf{Con} & \textbf{Coh} & \textbf{Flu} \\
    \midrule
    XSum & Fixed-buffer LLaDA & 15.55 & 38.4 & 0.285 & 0.218 & 0.883 & 3.29 & 3.96 & 3.35 & 4.54 \\
    XSum & \textsc{Delta}+LLaDA & 2.3 & 99.6 & 0.286 & 0.217 & 0.882 & 3.33 & 3.96 & 3.38 & 4.53 \\
    XSum & Fixed-buffer Dream & 15.74 & 38.1 & 0.334 & 0.259 & 0.892 & 2.81 & 3.81 & 3.09 & 4.46 \\
    XSum & \textsc{Delta}+Dream & 2.3 & 99.7 & 0.333 & 0.255 & 0.890 & 2.91 & 3.83 & 3.18 & 4.47 \\
    \midrule
    CNN/DM & Fixed-buffer LLaDA & 27.35 & 23.9 & 0.423 & 0.282 & 0.878 & 4.21 & 4.69 & 3.97 & 4.86 \\
    CNN/DM & \textsc{Delta}+LLaDA & 2.4 & 99.6 & 0.363 & 0.259 & 0.876 & 4.19 & 4.66 & 3.94 & 4.84 \\
    CNN/DM & Fixed-buffer Dream & 26.20 & 24.6 & 0.433 & 0.288 & 0.881 & 4.02 & 4.71 & 3.95 & 4.89 \\
    CNN/DM & \textsc{Delta}+Dream & 3.0 & 99.3 & 0.409 & 0.281 & 0.880 & 4.02 & 4.66 & 3.92 & 4.86 \\
    \midrule
    arXiv & Fixed-buffer LLaDA & 113.94 & 4.4 & 0.448 & 0.235 & 0.838 & 4.72 & 4.76 & 4.39 & 4.89 \\
    arXiv & \textsc{Delta}+LLaDA & 2.8 & 98.4 & 0.443 & 0.234 & 0.837 & 4.65 & 4.69 & 4.31 & 4.89 \\
    arXiv & Fixed-buffer Dream & 92.31 & 8.1 & 0.432 & 0.238 & 0.854 & 3.93 & 4.29 & 3.64 & 4.47 \\
    arXiv & \textsc{Delta}+Dream & 3.1 & 98.3 & 0.431 & 0.236 & 0.853 & 3.94 & 4.28 & 3.62 & 4.45 \\
    \bottomrule
  \end{tabular}
  \caption{Fixed-buffer ablation. Disabling \textsc{Delta}'s insert/remove operations substantially reduces character-length compliance while leaving summary quality broadly similar. G-Eval scores use GPT-4o-mini.}
  \label{tab:fixed_buffer}
\end{table*}

\section{DAEDAL Comparison}
\label{sec:daedal}

DAEDAL targets open-ended variable-length diffusion generation rather than generation toward a specified decoded character budget. We nevertheless evaluate DAEDAL with LLaDA using the same character-target prompts as the other generative methods. 

\begin{table*}[t]
  \centering
  \footnotesize
  \setlength{\tabcolsep}{3.5pt}
  \begin{tabular*}{\textwidth}{@{\extracolsep{\fill}}llrrrrrrrrrrr@{}}
    \toprule
    \textbf{Dataset} & \textbf{Method} & \textbf{$n$} & \textbf{MAE} & \textbf{$<$10} &
    \textbf{R-1} & \textbf{R-L} & \textbf{BF1} &
    \textbf{Rel} & \textbf{Con} & \textbf{Coh} & \textbf{Flu} & \textbf{Time} \\
    \midrule
    XSum & DAEDAL (LLaDA) & 11,334 & 74.5 & 7.7 & 0.251 & 0.196 & 0.845 & 2.82 & 3.41 & 2.81 & 4.25 & 2.50 \\
    XSum & \textsc{Delta}+LLaDA & 11,334 & 2.3 & 99.6 & 0.286 & 0.217 & 0.882 & 3.33 & 3.96 & 3.38 & 4.53 & 0.69 \\
    \midrule
    CNN/DM & DAEDAL (LLaDA) & 11,490 & 192.3 & 3.4 & 0.367 & 0.248 & 0.858 & 4.00 & 4.43 & 3.67 & 4.69 & 5.94 \\
    CNN/DM & \textsc{Delta}+LLaDA & 11,490 & 2.4 & 99.6 & 0.363 & 0.259 & 0.876 & 3.73 & 4.40 & 3.64 & 4.70 & 8.59 \\
    \midrule
    arXiv & DAEDAL (LLaDA) & 100 & 6337.5 & 1.0 & 0.106 & 0.073 & 0.796 & 2.64 & 2.92 & 2.34 & 2.90 & 306.56 \\
    arXiv & \textsc{Delta}+LLaDA & 6,440 & 2.8 & 98.4 & 0.443 & 0.234 & 0.837 & 4.65 & 4.69 & 4.31 & 4.89 & 22.80 \\
    \bottomrule
  \end{tabular*}
  \caption{Comparison with DAEDAL. Time is wall-clock seconds per summary. DAEDAL on arXiv is evaluated on a 100-example diagnostic subset because of its high runtime cost; \textsc{Delta}+LLaDA uses the full test set. G-Eval scores use GPT-4o-mini.}
  \label{tab:daedal}
\end{table*}

The full comparison is reported in Table~\ref{tab:daedal}. DAEDAL produces reasonable summaries but does not reliably satisfy specified character budgets. Across all three datasets, \textsc{Delta} reduces character-length error by a large margin, supporting the need for explicit decoded character-length estimation and correction.

\section{Runtime Analysis}
\label{sec:runtime}

Runtime results are reported in Table~\ref{tab:runtime}.

\begin{table}[H]
  \centering
  \resizebox{\columnwidth}{!}{%
  \begin{tabular}{lrrr}
    \toprule
    \textbf{Method} & \textbf{XSum} & \textbf{CNN/DM} & \textbf{arXiv} \\
    \midrule
    \textsc{Delta}+LLaDA & $0.69 \pm 0.43$ & $8.59 \pm 7.91$ & $22.80 \pm 22.98$ \\
    \textsc{Delta}+Dream & $0.67 \pm 0.39$ & $8.59 \pm 7.98$ & $20.75 \pm 21.72$ \\
    LA-SF-AR & $1.43 \pm 0.33$ & $2.19 \pm 0.64$ & $13.22 \pm 8.04$ \\
    DAEDAL (LLaDA) & $2.50 \pm 7.54$ & $5.94 \pm 4.22$ & $306.56 \pm 233.80^{*}$ \\
    BART-CNN & -- & $0.37 \pm 0.06$ & -- \\
    Lead-char-snap & $<0.01$ & $<0.01$ & $<0.01$ \\
    TextRank & $<0.01$ & $<0.01$ & $<0.01$ \\
    \bottomrule
  \end{tabular}%
  }
  \caption{Wall-clock runtime in seconds per summary (mean $\pm$ standard deviation). $^{*}$DAEDAL on arXiv is measured on a 100-example diagnostic subset because of its high runtime cost. GPT-4o-mini is omitted because concurrent API calls make per-sample latency not directly comparable.}
  \label{tab:runtime}
\end{table}

\section{Lead Baseline Diagnostic} \label{sec:lead3} 

To distinguish our character-budgeted Lead-char-snap baseline from the conventional Lead-3 baseline, we additionally evaluate standard Lead-3 on CNN/DailyMail. Table~\ref{tab:lead3} shows that Lead-3 achieves stronger extractive-summary quality but substantially violates the target character budget.

\begin{table*}[h]
\centering \small \begin{tabular}{lrrrrrrrrrr} \toprule \textbf{Method} & \textbf{Mean chars} & \textbf{MAE} & \textbf{$<$10} & \textbf{R-1} & \textbf{R-L} & \textbf{BF1} & \textbf{Rel} & \textbf{Con} & \textbf{Coh} & \textbf{Flu} \\ \midrule Lead-char-snap & 252.5 & 59.4 & 21.4 & 0.363 & 0.230 & 0.868 & 2.85 & 4.92 & 4.22 & 4.40 \\ Standard Lead-3 & 455.4 & 174.8 & 2.6 & 0.400 & 0.250 & 0.869 & 3.51 & 4.96 & 4.67 & 4.75 \\ \textsc{Delta}+Dream & 311.1 & 3.0 & 99.3 & 0.409 & 0.281 & 0.880 & 3.97 & 4.72 & 4.75 & 4.73 \\ \bottomrule \end{tabular} \caption{Standard Lead-3 diagnostic on CNN/DailyMail. Lead-3 achieves stronger extractive-summary quality than Lead-char-snap but substantially violates the target character budget. G-Eval scores use GPT-5-mini.} \label{tab:lead3} \end{table*} 

\section{LLM Judge Comparison}
\label{sec:judge_comparison}

We compare the original GPT-4o-mini G-Eval scores with GPT-5-mini. The two judges differ substantially on consistency and coherence, particularly for extractive baselines. GPT-5-mini assigns substantially higher consistency scores to Lead-char-snap, which is more consistent with its extractive nature. We therefore use GPT-5-mini as the primary judge in the main paper; the complete comparison is reported in Table~\ref{tab:judge_comparison}.

\begin{table*}[!t]
\centering
  \small
  \begin{tabular}{lllrrrr}
    \toprule
    \textbf{Dataset} & \textbf{Method} & \textbf{Judge} &
    \textbf{Rel} & \textbf{Con} & \textbf{Coh} & \textbf{Flu} \\
    \midrule
    XSum & \textsc{Delta}+LLaDA & GPT-4o-mini & 3.33 & 3.98 & 3.38 & 4.54 \\
         &                         & GPT-5-mini  & 3.41 & 4.21 & 4.61 & 4.66 \\
    XSum & \textsc{Delta}+Dream & GPT-4o-mini & 2.90 & 3.84 & 3.18 & 4.48 \\
         &                        & GPT-5-mini  & 3.17 & 4.14 & 4.76 & 4.82 \\
    XSum & LA-SF-AR & GPT-4o-mini & 3.49 & 4.02 & 3.55 & 4.64 \\
         &          & GPT-5-mini  & 3.56 & 4.38 & 4.79 & 4.89 \\
    XSum & GPT-4o-mini & GPT-4o-mini & 3.85 & 4.33 & 3.87 & 4.85 \\
         &             & GPT-5-mini  & 3.83 & 4.73 & 4.96 & 4.99 \\
    XSum & Lead-char-snap & GPT-4o-mini & 1.95 & 3.54 & 1.77 & 3.94 \\
         &                & GPT-5-mini  & 1.80 & 4.72 & 3.08 & 3.31 \\
    XSum & TextRank & GPT-4o-mini & 1.70 & 3.69 & 1.69 & 4.15 \\
         &          & GPT-5-mini  & 1.74 & 4.79 & 4.07 & 4.59 \\
    \midrule
    CNN/DM & \textsc{Delta}+LLaDA & GPT-4o-mini & 4.19 & 4.66 & 3.94 & 4.84 \\
           &                         & GPT-5-mini  & 4.04 & 4.59 & 4.58 & 4.49 \\
    CNN/DM & \textsc{Delta}+Dream & GPT-4o-mini & 4.00 & 4.65 & 3.91 & 4.85 \\
           &                        & GPT-5-mini  & 3.97 & 4.72 & 4.75 & 4.73 \\
    CNN/DM & LA-SF-AR & GPT-4o-mini & 4.48 & 4.75 & 4.14 & 4.97 \\
           &          & GPT-5-mini  & 4.29 & 4.61 & 4.94 & 4.96 \\
    CNN/DM & GPT-4o-mini & GPT-4o-mini & 4.79 & 4.89 & 4.34 & 4.98 \\
           &             & GPT-5-mini  & 4.53 & 4.86 & 4.99 & 5.00 \\
    CNN/DM & Lead-char-snap & GPT-4o-mini & 3.00 & 4.42 & 3.07 & 4.50 \\
           &                & GPT-5-mini  & 2.85 & 4.92 & 4.22 & 4.40 \\
    CNN/DM & TextRank & GPT-4o-mini & 2.64 & 4.08 & 2.67 & 4.04 \\
           &          & GPT-5-mini  & 2.40 & 4.80 & 3.38 & 4.07 \\
    CNN/DM & BART-CNN & GPT-4o-mini & 3.84 & 4.63 & 3.78 & 4.74 \\
           &          & GPT-5-mini  & 3.69 & 4.83 & 4.60 & 4.61 \\
    \midrule
    arXiv & \textsc{Delta}+LLaDA & GPT-4o-mini & 4.64 & 4.69 & 4.33 & 4.89 \\
          &                         & GPT-5-mini  & 3.71 & 4.35 & 4.43 & 4.39 \\
    arXiv & \textsc{Delta}+Dream & GPT-4o-mini & 3.94 & 4.27 & 3.63 & 4.44 \\
          &                        & GPT-5-mini  & 3.09 & 4.42 & 3.60 & 3.78 \\
    arXiv & LA-SF-AR & GPT-4o-mini & 4.87 & 4.83 & 4.61 & 4.98 \\
          &          & GPT-5-mini  & 3.98 & 4.28 & 4.78 & 4.98 \\
    arXiv & GPT-4o-mini & GPT-4o-mini & 4.98 & 4.96 & 4.91 & 5.00 \\
          &             & GPT-5-mini  & 4.35 & 4.65 & 4.97 & 5.00 \\
    arXiv & Lead-char-snap & GPT-4o-mini & 3.98 & 4.76 & 3.78 & 4.58 \\
          &                & GPT-5-mini  & 2.22 & 4.99 & 4.11 & 4.27 \\
    arXiv & TextRank & GPT-4o-mini & 3.23 & 3.88 & 3.00 & 3.91 \\
          &          & GPT-5-mini  & 2.52 & 4.55 & 3.00 & 3.19 \\
    \bottomrule
  \end{tabular}
  \caption{Comparison of GPT-4o-mini and GPT-5-mini as G-Eval judges.}
  \label{tab:judge_comparison}
\end{table*}

\section{Summarization Prompts}
\label{sec:prompts}

We use dataset-specific prompt templates for all generative methods (\textsc{Delta}, LA-SF-AR, GPT-4o-mini, and DAEDAL). The \texttt{\{target\_len\}} placeholder is replaced with the target character length and \texttt{\{source\}} with the source document.

\paragraph{CNN/DailyMail.}
\begin{quote}
\small\ttfamily
Summarize the following article in approximately \{target\_len\} characters:\textbackslash n\textbackslash n\{source\}
\end{quote}

\paragraph{arXiv.}
\begin{quote}
\small\ttfamily
Write an abstract for the following scientific paper in approximately \{target\_len\} characters:\textbackslash n\textbackslash n\{source\}
\end{quote}

\paragraph{XSum.}
\begin{quote}
\small\ttfamily
Write a one-sentence summary of the following article in approximately \{target\_len\} characters:\textbackslash n\textbackslash n\{source\}
\end{quote}

\subsection{G-Eval Evaluation Prompt}

We use GPT-5-mini as the primary judge with the following system and user prompts. Scores range from 1 (poor) to 5 (excellent). We additionally evaluate the same outputs with GPT-4o-mini for the judge comparison reported in Appendix~\ref{sec:judge_comparison}.

\paragraph{System prompt.}
\begin{quote}
\small\ttfamily
You are an expert evaluator tasked with assessing the quality of text summaries.

Evaluate the provided summary against the source text based on the following criteria:

- Relevance: Does the summary capture only the most important information from the source text?

- Consistency: Is the summary factually consistent with the source text?

- Coherence: Is the summary well-structured and logically organized?

- Fluency: Is the summary written in clear, grammatical English?

Score each criterion on a scale from 1 to 5:

1: Poor --- Fails to meet the criterion

2: Below Average

3: Average

4: Good

5: Excellent --- Perfectly meets the criterion

Respond only in the following JSON format, where X is your score:

\{"relevance": X, "consistency": X, "coherence": X, "fluency": X\}
\end{quote}

\paragraph{User prompt.}
\begin{quote}
\small\ttfamily
Evaluate the following summary against the source text based on the criteria defined above.

[Source Text]: '\{source\}'

[Summary]: '\{summary\}'

Respond only in the following JSON format, where X is your integer score for each criterion:

\{"relevance": X, "consistency": X, "coherence": X, "fluency": X\}
\end{quote}

\balance
\section{Implementation Details}
\label{sec:baseline_details}

\paragraph{TextRank.} TF-IDF sentence scoring \citep{mihalcea2004textrank}, truncated to the target character length at the nearest sentence boundary.

\paragraph{Lead-char-snap.} Selects the maximum number of initial source sentences that fit within the target character budget. This avoids unnatural mid-sentence truncation while providing an extractive baseline with approximate character-length control.

\paragraph{BART-CNN.} BART-large fine-tuned on CNN/DailyMail \citep{lewis2020bart}, beam size 4, length penalty 2.0.

\paragraph{GPT-4o-mini.} Prompted with the target character count using the templates in Appendix~\ref{sec:prompts}.

\paragraph{LA-SF-AR.} The zero-shot pipeline of \citet{retkowski2025zeroshot} using Llama-3.1-8B-Instruct with length approximation ($\mu_w{=}6.31$ chars/word), sample filtering ($N{=}3$, temperature 0.7), and automated revisions (up to 3, $\epsilon{=}10\%$).

\paragraph{DAEDAL.} We evaluate DAEDAL \citep{li2026beyond} with LLaDA using the same dataset-specific character-target prompts as the other generative methods. On arXiv, we evaluate DAEDAL on a 100-example diagnostic subset because of its high inference cost.

\paragraph{\textsc{Delta} hyperparameters.} On CNN/DailyMail and XSum: $(\tau, \tau_{\text{low}}) = (0.65, 0.95)$ for Dream, $(0.65, 0.90)$ for LLaDA. On arXiv: $(0.90, 0.82)$ for Dream, $(0.71, 0.71)$ for LLaDA. Selected via grid search on the validation set. The initial buffer size is determined using the average number of characters per token estimated from validation references. The maximum number of iterations is set to $\lfloor L^* / 2 \rfloor$.

\paragraph{Reproducibility.} We release our implementation, fixed-buffer ablation code, DAEDAL evaluation script, prompts, hyperparameters, random seeds, and evaluation scripts at \url{https://github.com/DFKI-NLP/delta-diffusion}.

\section{Use of AI Assistants}
\label{sec:ai_assistants}

Claude (Anthropic) was used to assist with writing and editing portions of the manuscript, generating figure components, and developing parts of the codebase. All outputs were reviewed and verified by the authors.

\end{document}